\documentclass[conference]{IEEEtran}
\IEEEoverridecommandlockouts
\usepackage{cite}
\usepackage{amsmath,amssymb,amsfonts}
\usepackage{algorithmic}
\usepackage{graphicx}
\usepackage{textcomp}
\usepackage{xcolor}
\usepackage{cite}
\usepackage{amsmath,amssymb,amsfonts}
\usepackage{algorithmic}
\usepackage{graphicx}
\usepackage{textcomp}
\usepackage{subcaption}
\usepackage{xcolor}
\usepackage{comment}
\usepackage[linesnumbered,ruled,vlined]{algorithm2e}% 
\usepackage[]{algorithm2e}
\usepackage[]{algorithmic}
\def\BibTeX{{\rm B\kern-.05em{\sc i\kern-.025em b}\kern-.08em
    T\kern-.1667em\lower.7ex\hbox{E}\kern-.125emX}}

\begin{document}

\title{A deep learning pipeline for identification of motor units in musculoskeletal ultrasound}

%\footnote{This work was supported by the Swedish research council (dnr 2015-04461) and the Kempe foundations (dnr JCK-1115 and SMK-1868)}

\author{\IEEEauthorblockN{Hazrat Ali\IEEEauthorrefmark{1}, Johannes Umander \IEEEauthorrefmark{1},
Robin Rohl{\'e}n\IEEEauthorrefmark{1}, 
and Christer Gr{\"o}nlund}\IEEEauthorrefmark{1}

\IEEEauthorblockA{\IEEEauthorrefmark{1}Department of Radiation Sciences, Umea University, Sweden. \\Email: hazrat.ali@umu.se, johannes.umander@ume.se, robin.rohlen@umu.se, christer.gronlund@umu.se}
\thanks{This work was supported by the Swedish research council (dnr 2015-04461) and the Kempe foundations (dnr JCK-1115 and SMK-1868)}.
\thanks{*Correspondence to: christer.gronlund@umu.se}
\thanks{This work is accepted for publication in IEEE Access.}
}

%\thanks{Manuscript received December 2018; revised xx xx, 2018.}
%}
%
%\markboth{IEEE Access Journal,~Vol.~xx, No.~xx, December~20xx}%
%{Kashif \MakeLowercase{\textit{et al.}}: A spatio-temporal analytics approach for mobile traffic optimization using call detail records}

\maketitle

\begin{abstract}
Skeletal muscles are functionally regulated by populations of so-called motor units (MUs). An MU comprises a bundle of muscle fibers controlled by a neuron from the spinal cord. Current methods to diagnose neuromuscular diseases and monitor rehabilitation, and study sports sciences rely on recording and analyzing the bio-electric activity of the MUs. However, these methods provide information from a limited part of a muscle. Ultrasound imaging provides information from a large part of the muscle. It has recently been shown that ultrafast ultrasound imaging can be used to record and analyze the mechanical response of individual MUs using blind source separation. In this work, we present an alternative method - a deep learning pipeline - to identify active MUs in ultrasound image sequences, including segmentation of their territories and signal estimation of their mechanical responses (twitch train). We train and evaluate the model using simulated data mimicking the complex activation pattern of tens of activated MUs with overlapping territories and partially synchronized activation patterns. Using a slow fusion approach (based on 3D CNNs), we transform the spatiotemporal image sequence data to 2D representations and apply a deep neural network architecture for segmentation. Next, we employ a second deep neural network architecture for signal estimation.
The results show that the proposed pipeline can effectively identify individual MUs, estimate their territories, and estimate their twitch train signal at low contraction forces. The framework can retain spatio-temporal consistencies and information of the mechanical response of MU activity even when the ultrasound image sequences are transformed into a 2D representation for compatibility with more traditional computer vision and image processing techniques. The proposed pipeline is potentially useful to identify simultaneously active MUs in whole muscles in ultrasound image sequences of voluntary skeletal muscle contractions at low force levels.
\end{abstract}


% \fi
\IEEEpeerreviewmaketitle

\section{Introduction}
The motor unit (MU) is the smallest voluntarily activatable unit in the skeletal muscles. Its function (and control) is important to study for the diagnosis of neuromuscular diseases and understanding of skeletal muscle physiology in sports sciences and rehabilitation \cite{sale1987}, \cite{henderson2017}. An MU is defined as a motor neuron connected to a bundle of muscle fibers located within a given territory (Fig. 1A) \cite{Basmajian1985}. The control of an MU is encoded in a firing pattern (Fig. 1A) originating in the spinal cord mediated by the motor neuron. The corresponding output is characterized by repeated electrical depolarizations of the fibers and subsequent repeated shortening and thickening of the fibers (mechanical twitch train) \cite{lind2018} (Fig. 1A and B). 

Electromyography (EMG) is the gold standard technique to study MUs where electrodes are used to record the fibers' electrical activity either invasively or from the surface of the skin \cite{king1997concentric, martinez2016high,farina2016characterization}. This technique provides high-quality data, but due to a low pass filtering effect of the tissue, there is a restricted field of view \cite{fuglevand1992detection}. Ultrasound imaging is a non-invasive technique allowing mechanical information from a large field of view \cite{szabo2004}.

Recently our group presented a method \cite{rohlen2020method} to study MU activity based on the mechanical response of individual MUs using ultrafast ultrasound imaging (>2000 images per second \cite{bercoff2011ultrafast}). This method was based on a blind source separation framework and decomposition of spatio-temporal components. However, the performance was found to decrease with an increasing number of active MUs in the contractions.
This problem's challenge is that muscle activation is a highly complex physiological process, where tens to hundreds of MUs with overlapping MU territories can be active simultaneously with individual firing patterns (Fig. 1A and C).

One interesting approach that has shown tremendous potential to learn complex patterns is using deep learning models comprising neural networks with several layers' architecture. In particular, in medical imaging, there is a vast literature on deep learning applications for detection and segmentation \cite{dlmedicalreview}. \textit{In this work, we hypothesize that a deep learning methodology can improve the performance of identification of individual MUs, by learning the underlying complex interaction patterns using the full image sequence information.}

This work aims to develop and evaluate a deep learning pipeline to 1) detect individual active MUs, 2) segment their territories, and 3) estimate their activation twitch signal, using ultrafast image sequences of voluntary skeletal muscle contractions. The model is trained and evaluated using simulated ultrasound data \cite{rohlen2020method}.

This work presents a deep learning-based method to identify individual MUs in spatio-temporal data of contracting muscles. To the best of our knowledge, it is the first deep learning approach to identify a varying number of fixed-position objects with unique individual temporal patterns of intensity changes.

The rest of the paper is organized as follows: In Section II, we review related work. In Section III, we present the proposed deep learning pipeline in detail and also present the evaluation metrics. Section IV gives an overview of the simulation model and data sets generated. Section V presents the results, and Section VI gives a discussion of the findings. Finally, Section VII concludes the paper.

\begin{figure*}[ht!]
\begin{center}
%\makebox[\textwidth]
{\includegraphics[width=0.9\textwidth]{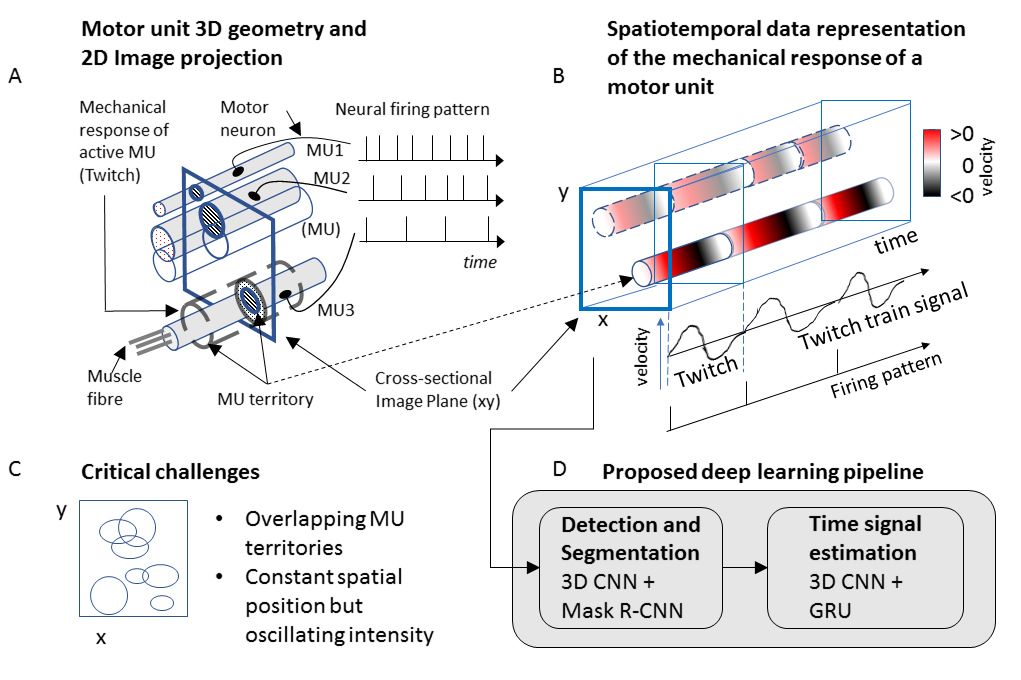}}
\end{center}
\caption{Illustration of principal skeletal muscle anatomy and recorded ultrasound image sequence. A) Shows three motor units, aligned in parallel, and they are activated with unique neural firing patterns. An activated unit's mechanical response results in a twitch - a thickening and shortening of the unit's fibers. B) An illustration of the spatio-temporal features of the recorded image sequence from the cross-sectional plane. C) The main challenges of the data, including the overlap of motor unit territories. D) An overview of the proposed deep learning approach comprising two modules of detection and segmentation and time signal detection.
}
\label{fig:overview_method}
\end{figure*}

\section{Related work}
Deep learning \cite{lecun2015deep, schmidhuber2015deep, DLbook} has greatly revolutionized many different domains involving analysis of a large image, audio, text, video, or tabular data. Of particular relevance to the work reported in this paper are the advancement made in image and video processing using deep learning methods for segmentation, identification, recognition \cite{dlvideosegreview}, deep learning for time-series data (e.g., speech) \cite{graves2013speech}, and deep learning in medical imaging \cite{dlmedicalreview, migan2018}. Hence, we present the relevant details on deep learning for segmentation and signal detection in the subsequent text.

\subsection{Object detection and segmentation}
The advancements made on instance segmentation tasks in the computer vision field have paved the way for many progress applications for deep learning applications in the medical imaging domain. Traditionally, the classification task was performed to categorize an image into a distinct class (e.g., cat vs. dog classification). For a more realistic task, we are usually interested in the object's position within an image, referred to as the localization task. When there are multiple instances of the same object within an image, we perform object detection, which localizes the object. For more practical use, we do pixel level localization of the object referred to as the segmentation. 

In semantic segmentation, each class in the image is masked with a different color. For example, all the pixels containing dogs might be colored blue, and all the pixels containing cats may be colored red. A problem with this approach is when objects of the same class overlap, they are merged under the same mask, and it is difficult to differentiate between them. To solve this problem, one can use instance segmentation where every object (instance) gets its own mask \cite{rcnn, fastrcnn, fasterrcnn, maskrcnn}.

One of the simplest methods to perform object detection is to crop out multiple locations of an image and run a Convolutional Neural Network (CNN) to classify the cropped region. The problem with this approach is that it is extremely slow. R-CNN (Region-based CNN) \cite{rcnn} tried to solve this by first applying a non-learning-based algorithm called a selective search that returned 2,000 likely region proposals that a CNN then classified and predicted a more refined bounding box around the object. This approach is considered slow because CNN has to classify 2,000 regions. Fast R-CNN \cite{fastrcnn} solved this problem by using a CNN to process the entire image into a convolutional features map. Then, to classify one of the region proposals, one can crop out a region of this features map corresponding to the region in the image using RoI pooling and then classify that data. This approach is faster than R-CNN, but it still relies on the non-learning based selective search to find interesting region proposals. Faster R-CNN\cite{fasterrcnn} replaced the selective search algorithm with a new network called region proposal network (RPN) that used the information from the convolutional features map to generate region proposals. 

Mask R-CNN\cite{maskrcnn} is an extension of the Faster R-CNN architecture to introduce instance segmentation. To achieve this, mask R-CNN modified some parts of the network. The convolutional features map is replaced by a feature pyramid network (FPN), which contains feature maps at multiple scales of the image. RoI pooling is replaced with a new method called RoIAlign, which works better when pixel-level accuracy is required. A pixel map containing the object is generated by adding a new head to the classifier and bounding box predictor heads for predicting the mask.

\subsection{Time signal estimation}
Typically, recurrent neural networks (RNN) have been popular with time-series data processing. Successful use cases have been reported for sequential data, in particular in natural language processing. Comparative studies have shown the benefit of RNN for time dependencies modeling and signal tracking \cite{graves2013speech, yin2017comparative, ugurlu2018electricity, wang2018disconnected, greff2016lstm}. For example, the work in \cite{graves2013speech} achieved a state-of-the-art error reduction on the popular TIMIT dataset for speech recognition. The Gate Recurrent Unit (GRU) is a type of RNN that helps overcome the vanishing gradient issue in training an RNN. This issue is typically achieved by updating and reset gates controlling the inward and outward flow of information from the neural network's memory states. This process effectively helps in removing information that is redundant for the prediction task. For additional details on RNN and GRU, we refer the reader to \cite{graves2013speech, yu2019review}.

\section{Proposed Model}

\subsection{Prerequisites and overview}
Three key features characterize the mechanical response of individual MUs that we want to identify:
\begin{enumerate}
\item {Spatio-temporal characteristics of units:} The mechanical response of an MU is characterized by a fixed location of a spatial territory with varying intensity (Fig 1B). The time signal intensity variation of a unit is unique due to an MU's unique neural firing pattern (Fig 1A) \cite{Basmajian1985}.
\item {Unknown number of units:} The number of active (and thus visible) MUs in an image sequence is unknown because the recruitment of units is highly complex coordination by the central nervous system \cite{Basmajian1985,rohlen2020method}.
\item {Overlapping territories of the units:} The territories of two or more MUs may be overlapping \cite{Basmajian1985}, causing spatial and temporal interference of their activity (Fig 1A and C).
\end{enumerate}

NB: In the text, we sometimes write \textit{object} for MU or unit, and \textit{signal estimation} for extraction of the mechanical twitch train signal, to harmonize with the terminology in machine learning and signal/image processing.

Given the spatio-temporal nature of the mechanical response of an MU (Fig 1B), we split the problem into two main modules (Fig 1D).

The first module is the \textbf{detection and segmentation model}, which detects and segments the MU territories within the image sequence. The second module, called the \textbf{time signal estimation model}, determines the mechanical activation signal (the twitch train) caused by a specified MU. As typical, the best parameters for the deep networks used here are determined empirically and through a grid-search on a finite set of parameters.

\subsection{Detection and segmentation model}
This module processes an image sequence to perform the detection and segmentation of the MUs. This process is particularly challenging due to the spatio-temporal nature of the mechanical response of MUs (as previously pointed out). 
In short, we use a 3D CNN, which helps to retain the temporal information while generating a 2D representation. The transformed 2D representation is used for instance segmentation using Mask R-CNN approach. 

\subsubsection{Architecture}
\begin{figure*}[ht]
\begin{center}
\makebox[\textwidth]{\includegraphics[width=\textwidth]{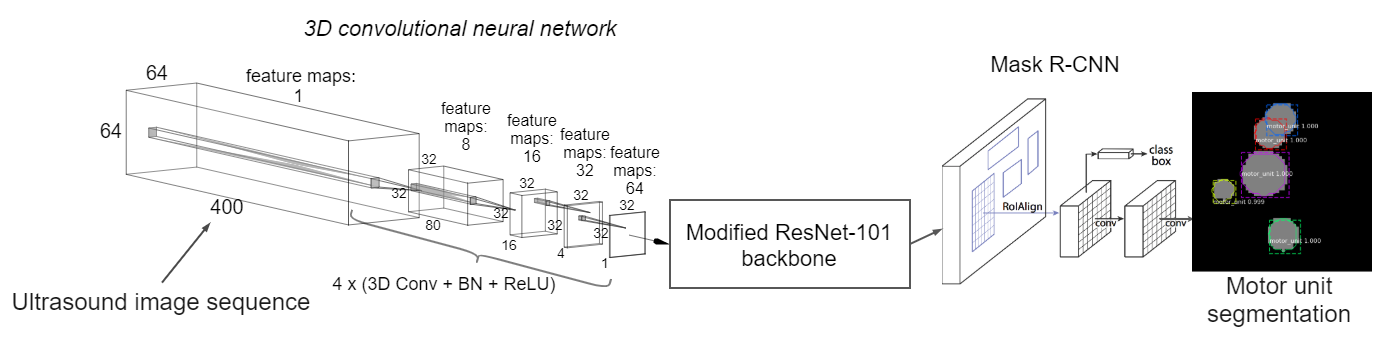}}
\end{center}
\caption{Detection and segmentation model architecture. Input data is the spatio-temporal image sequence of the complex activity of activated motor units in a skeletal muscle ($64 \times 64 \times 400$ is $Depth \times Width \times Frames$ and corresponds to 40$ \times $40 mm and 1 second). The output is the segmented spatial territories of identified motor units (right).}
\label{fig:seg_arch}
\end{figure*}

The first thing that takes place in the model is to convert the data into a 2D representation using a series of 3D convolutions. The slow fusion approach \cite{slowfusion} inspires this way of processing image sequence data. The network preserves the temporal information on the action potential. More global information is made accessible to the higher layers in the network, retaining both the ultrasound sequence's spatial and temporal aspects. During this transformation, we keep the spatial resolution while reducing the temporal dimension. This architecture is visualized on the left part of Figure \ref{fig:seg_arch}.

The first layer receives the standardized image sequence of size $64\times64\times400\times1$, and it has to reduce the size of the data to make the computational problem feasible. It uses a strided convolution of dimensions $2\times2\times5$ (\textit{$height \times width \times  time$}) to reduce the number of computations required and also reduces the spatial dimensions by a factor of two and the temporal dimension by a factor of 5. It uses a kernel size of ($7\times7\times7$) and 8 feature maps followed by batch normalization (BN) and the ReLU activation function. The strided convolution reduces the size of the data to $32\times32\times80\times8$.\
A second convolution is performed on this data, which uses a kernel size of ($3\times3\times7$) and 16 feature maps followed by BN and ReLU. Since performance is not as critical at this point compared to the first layer due to the reduced data size, max pooling is used instead of strided convolutions to reduce the temporal dimension. A max pooling layer reduces the temporal dimension by a factor of 5 to produce a data size of $32\times32\times16\times16$.\
The third convolution uses a kernel size of ($3\times3\times5$) and 32 feature maps followed by BN, ReLU, and a max pooling layer, reducing the temporal dimension a factor of 4. This results in a data size of $32\times32\times4\times32$.\
At this stage, the final convolution uses a kernel of size $3\times3\times5$, 64 feature maps, BN, ReLU, and a max pooling layer that reduces the temporal dimension a factor of 4 to create a data size of $32\times32\times1\times64$.

At this stage, the data gets a 2D representation with $32 \times 32$ pixels and $64$ channels. We now employ a mask R-CNN model \cite{maskrcnn}, with implementation from  \cite{matterport_maskrcnn_2017} with a ResNet--101 model used for features extractions \cite{resnet} as the feature extractor \footnote{The implementation is in Python3 and uses the tensorflow and keras frameworks.} (See Fig. \ref{fig:seg_arch}).  
An issue that arises is that ResNet is normally used for high-resolution images. The ResNet network's first stage reduces the spatial dimensions by a factor of 4, which is unwanted when we already have a low-resolution image and would result in a very poor segmentation. In our work, this issue is addressed by removing the first stage of the ResNet network.

The actual object detection and segmentation is then performed by Mask R-CNN, which receives the features generated by ResNet-101 through a feature pyramid network (FPN). The Mask R-CNN network in this model retains the default parameters used in the implementation.

\subsubsection{Performance metrics}
We evaluate the detection and segmentation performances through precision and recall measures. 

An MU is considered correctly detected if the intersection over union (IoU) measure for the MU mask is greater than 0.5. The detection step drives the segmentation performance, as we would like to consider MUs that as classified as true positives during the detection step.

\subsubsection{Training process}
Ten thousand (10,000) simulated images sequences were generated for the training (see section IV. B Datasets). To further increase the training data, we perform data augmentation (random flipping) and get a total training set of 40,000 simulated image sequences, containing 699,788 MUs.

To speed up the training process and achieve better convergence, we used transfer learning. 
We use ResNet-101 \cite{resnet101} and Mask R-CNN models \cite{maskrcnn}, pre-trained on the Microsoft COCO dataset \cite{coco2014}. The following changes are introduced. The COCO dataset has 80 different categories. So, the classifier layer is modified to suit our task of binary classification. The 3D CNN layers are trained from scratch with weights initialized using the Xavier initialization method \cite{xavier2010}.

The model is trained using stochastic gradient descent (SGD) with an initial learning rate of 0.001 and momentum set to 0.9 \footnote{The model is trained on an Nvidia RTX 2070 with 8GB of memory which allowed for 8 examples per mini-batch.}. 

For the first 20,000 mini-batches, the ResNet backbone weights are kept fixed so that to preserve the knowledge from the previous application and the other layers have to adjust to it. Afterward, the learning rate is reduced by a factor of 10 for training up to 200,000 mini-batches. Finally, the learning rate is further reduced by a factor of 10 as the training continues to the 500,000 mini-batch.  
The training takes a total of 67 hours as we get the optimal weights. The training and validation loss graphs can be seen in Figure \ref{fig:seg_losses} (in Appendix \ref{appendixA}). The loss function used for training this model is the sum of five different loss functions: the region proposal networks loss, the bounding box loss, the Mask R-CNNs loss, bounding box loss, and the mask loss.

In addition, to improve the model's tolerance to noise, the model is also further trained for an additional 100,000 mini-batches with a random noise level value between 30 to 10 dB SNR.

\subsection{Time signal estimation model}
The time signal estimation model is designed to learn to estimate the MU twitch train signal from a detected and segmented MU from the first model. Thus it removes signals that do not originate from the MU of interest. In short, with motivation from the slow fusion approach by Karpathy et al. \cite{slowfusion}, we first use a 3D CNN to transform the image sequences into a time-series data and then train a GRU network to estimate the twitch train.

\subsubsection{Architecture}
The input to the model is a cropped version of the spatio-temporal image sequence. It has a spatial resolution of $16\times16$ pixels and 400 timesteps (corresponding to a one-second sequence within a 10$\times$10 mm ROI), which is sufficient to encompass an MU size in the biceps muscles \cite{staalberg1991scanning}). 

Unlike the segmentation module, here we want to reduce the spatial dimensions and retain the temporal dimension. Similar to the segmentation module, a 3D CNN, inspired by the slow fusion concept \cite{slowfusion}, is adapted to transform the image sequences for subsequent time signal estimation. It is processed through five layers of 3D convolutions to extract spatio-temporal features.

The first convolution has a stride of 2 in two dimensions (say, $x$ and $y$) to reduce the spatial resolution to $8\times8$ and thus reduce the computations. 

\begin{figure*}[ht!]
\begin{center}
\makebox[.9\textwidth]{\includegraphics[width=\textwidth]{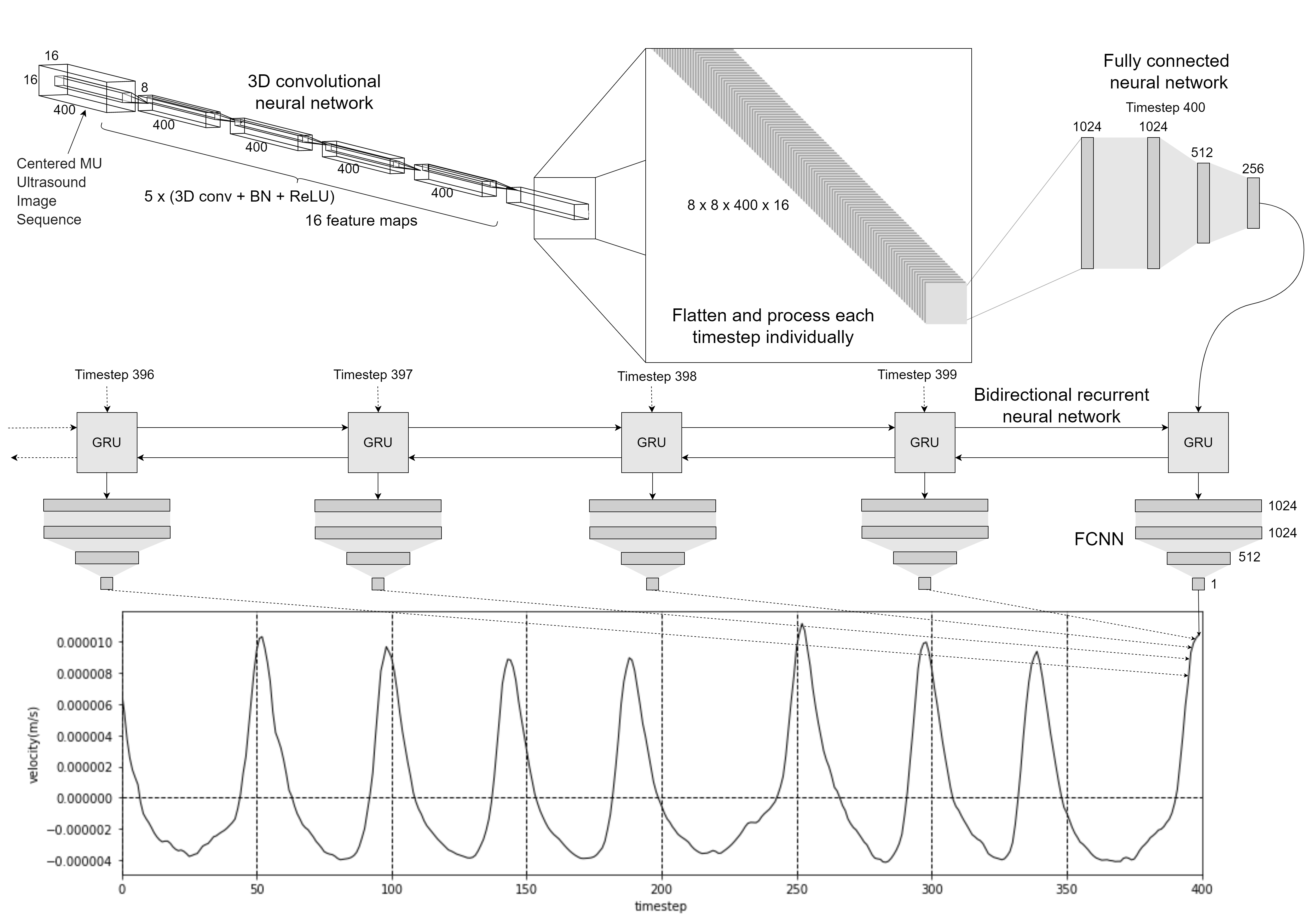}}
\end{center}
\caption{Time signal estimation model architecture. The input to this model is cropped image sequence data centered around the detected motor unit territory as segmented in model 1 ($16 \times 16 \times 400$ is $Depth \times Width \times Frames$ corresponding to 10 $\times$ 10 mm, and 1 second sequence). The output is the corresponding motor unit twitch train signal (bottom).}
\label{fig:transc_arch}
\end{figure*}

Each convolution layer creates 16 filters and uses a $3\times3\times15$ kernel except the first layer, which has a $5\times5\times15$ kernel.
Each layer is followed by batch normalization and then the ReLU activation function. The resulting data from these 3D convolutions is of dimension $8\times8\times400\times16$. The 3D CNN is visualized in the top left of Figure \ref{fig:transc_arch}.
This data is then sliced by the time steps and flattened so that each time step can be processed individually. The idea behind processing each time step individually is to create higher-level features that are spatially invariant, e.g., the width of the MU at that timestep. This processing is done in a fully connected neural network (FCNN). First, a layer of 1,024 neurons is applied, followed by a layer with 512 neurons, and finally, a layer with 256 neurons. The fully connected layers use the ReLU activation function, and all the time steps share the weights for the FCNN.

The data now consists of a feature vector of 256 numbers for each time step passed to two RNNs, each going in opposite directions, forming a bidirectional RNN. The motivation of using RNN here is to create a temporal signal based on the information from all the timesteps. For example, suppose the signal from an MU is unintelligible during an interval. In that case, RNNs could be used to look at the previous and preceding time steps to estimate the signal in the missing interval. Each RNN consists of 512 gated recurrent units (GRU). The RNNs are configured to return the entire sequence and not just the last output. The output of the RNNs is then concatenated, resulting in 1,024 values for each timestep.

The data is run through a final FCNN at each time step to produce the final result (the amplitude of the twitch train signal of the MU at each time step). First, a layer of 1,024 neurons is applied, followed by a layer with 512 neurons, and finally, a layer produces a single value (the amplitude). All layers except the final layer use ReLU, and all time steps share the weights. The data now consists of 400 values representing the mechanical response (tissue velocity) at each time step.

\subsubsection{Performance metrics}
We evaluate the extracted signals' performance by comparing the firing patterns of the estimated and simulated signals. The performance metric selected for this task is the rate of agreement (RoA), as it provides easy interpretation,  captures the information we are interested in \cite{RoA2016}, and was used in previous work on similar signals \cite{rohlen2020method}.  
The RoA measure ranges between 0 and 1 where 1 is a perfect score, and it measures the agreement between the firings of two signals. The formula for RoA is:
\begin{align}
   RoA = \frac{c}{c + A + B} , \;\; \text{such that } 0 \leq RoA \leq 1 
\end{align}

Here, $c$ is the number of times when both the true firings and the estimated firings have matched. $A$ is the number of times signal A had a firing but not signal B, and $B$ is the number of times signal B had a firing but not signal A. Two firings are considered matched if they occur within 15 ms (6 timesteps at 400 Hz) of each other. An example of how RoA is calculated for a given MU firing can be seen in Figure \ref{fig:RoA_tutorial} in Appendix \ref{appendixC}.

The MU firings are extracted as the local maxima of the estimated twitch train signals (Fig. \ref{fig:RoA_tutorial}). Before this, a low-pass filter is applied to the signals using a running average with a window of 11 timesteps (27.5 ms). Local maxima with an amplitude (velocity) of less than 0 m/s were excluded.

\subsubsection{Training process}

The twitch train estimation model uses the same training set as the segmentation model.
The model is trained for 100,000 mini-batches using Adam optimizer with a learning rate of 0.0001 and clipping the gradient norm to a maximum of 1 \footnote{8GB of video memory on the RTX 2070 allows for 32 simulations per mini-batch.}.

This model tends to quickly overfit the training data, as shown in Figure \ref{fig:transc_losses}. No regularization technique seems to fix this problem without also regressing the validation performance. Therefore, early stopping is used to choose the best performing model (on the validation set) instead of the last model. The training process took around 17 hours (although, as per Figure \ref{fig:transc_losses} shown in Appendix \ref{appendixB}, only about 10 hours were necessary to find the best performing model).

The loss function used for this model's training is the mean squared error between the predicted signal and the ground truth signal. However, one issue found with this loss function is that ground truth signals with large amplitudes tend to get much greater losses than ground truth signals with smaller amplitudes. This issue happens even though the predicted signals for the larger amplitude ground truth visually followed the signal much better. This issue results in that the optimizer primarily focusing on improving signals with larger amplitudes and ignoring MUs with weaker signals.
This problem was solved by normalizing all the ground truth signals to an amplitude of 1 when calculating the loss and scaling the predicted signals using the same coefficient.

\section{Simulation of muscle activation}
A simulator \cite{rohlen2020method} generated the data used for training and evaluating the models in this work. 
Since the simulator knows all the latent variables used to generate the data, it can also provide the labels for the example in the form of masks of the cross-sectional territory encompassed by each MU and the mechanical twitch train signal for each MU.

\subsection{Simulation model}
The simulation model used in this work was previously described in Rohl{\'e}n et al. \cite{rohlen2020method}, and here we give a brief description. The model generates the tissue velocity image sequences of a contracting muscle based on a modified EMG simulation model \cite{farina2001novel}, were the electrical action potential responses are replaced by mechanical spatio-temporal twitch responses. The mechanical response, in the plane perpendicular to the fiber direction (cross-sectional), is modeled (Fig 1A). An MU territory is modeled as a circular region, and the corresponding mechanical twitch response is modeled using in vivo empirical MU tissue velocity waveform from electro-stimulation experiments \cite{deffieux2008assessment}, \cite{gronlund2013imaging}. It is assumed that the force is transmitted along the fiber direction only and that there is no mechanical connectivity between the fibers of different MUs. Parameters were set to simulate a biceps brachii muscle at weak isometric contraction levels. The firing patterns of the MUs had a firing rate (FR) in the range 8 and 13 Hz (randomly distributed) with an inter-pulse-interval variation of $\mathcal{N}(0,0.2/FR)$ \cite{Basmajian1985}, \cite{stock2016motor}. Synchronization of MU firings was simulated in the range of 0-10 \% and was computed as the percentage of MU firings synchronized with (firings of) other MUs \cite{yao2000motor, kirkwood1991cross}. The territories of the MUs were randomly located within the simulated muscle cross-section and had a diameter in the range 2.5mm to 10 mm (randomly distributed) \cite{staalberg1991scanning}.
    
\subsection{Datasets}
We generated three datasets - training, validation, and test sets - consisting of 10,000, 1,000, and 600 simulated image sequences. %The same distribution is used for both the segmentation and the signal estimation models.

The simulated (tissue velocity) image sequences were 64 $\times$ 64 $\times$ 400 pixels, corresponding to 40 mm $\times$ 40 mm $\times$ 1 s with spatial resolution of 0.625 mm/pixel, and 400 Hz frame rate. 
Each simulated image sequence in the training and validation sets contains between 5 and 30 motor units. In contrast, the test set contains 100 simulated image sequences of each of the following categories: 1, 5, 10, 15, 20, and 25 MUs. 

Gaussian white noise was added to the simulated signals at 10, 20, and $\infty$ dB.

\section{Results}
Figure \ref{fig:ImageSequence5MUs} and \ref{fig:ImageSequence15MUs} of the Appendix present examples of the image sequences of simulated mechanical response of skeletal muscle activity with 5 and 15 active MUs. Figure \ref{fig:Example5MUsTrueAndEstimated} and \ref{fig:Example15MUsTrueAndEstimated} show the true and estimated territories and twitch train signals for these corresponding datasets. It can be seen that the complexity of the mechanical response pattern increases with an increasing number of active MUs. For example, when 5 units were active, all MUs were detected, but when 15 units were active, two units failed to be detected.

\subsection{Motor unit detection and segmentation performance}
\begin{figure*}[ht]
\label{RoA}
\begin{center}
\makebox[\textwidth]{\includegraphics[width=\textwidth]{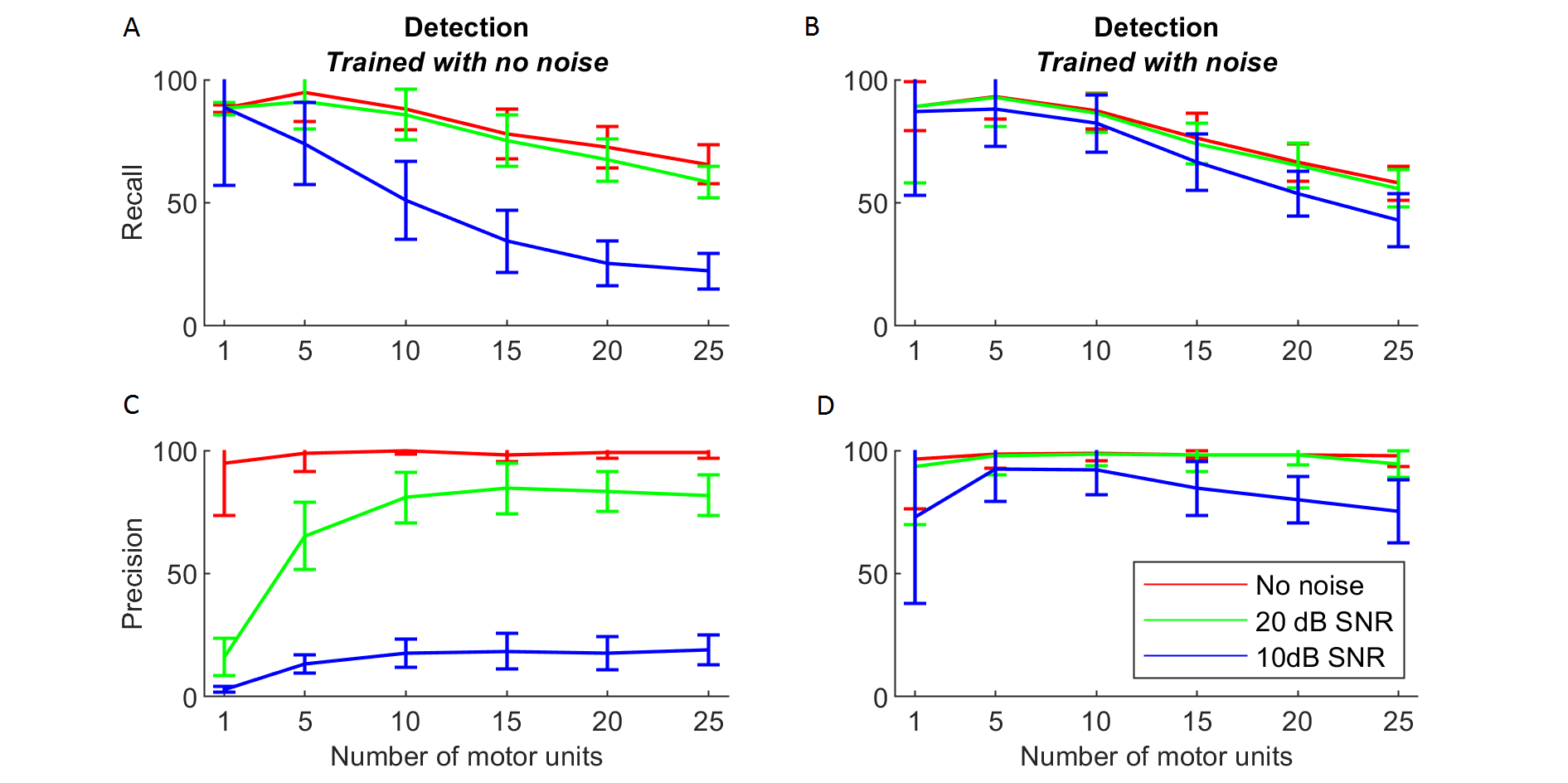}}
\end{center}
\caption{Object detection recall (A) and (B) and precision (C) and (D). The model's result was evaluated on the test set with a varying number of motor units and noise. Each data point shows the mean and standard deviation of 100 simulations. Figure \ref{fig:detect_results}(A) and (C) correspond to the model trained with noise-free data. Figure \ref{fig:detect_results} (B) and (D) correspond to the model trained on noisy data.}
\label{fig:detect_results}
\end{figure*}

The object detection results for the segmentation model trained, as described in the method section, can be seen in Figure \ref{fig:detect_results}. When the model is trained on signals without noise, it achieves high precision in the case when the data has no noise, which means that the model rarely makes miss predictions (i.e., lower false positives). When noise is applied at 20 dB SNR, the model still correctly detects most of the MUs, but it starts to fail to detect some MUs and detect some false MUs as can be seen in the reduced recall and precision, respectively. At 10dB SNR, the model performance is greatly reduced with respect to recall and precision.

In general, training the model with noisy data significantly improved the model (Fig.  \ref{fig:detect_results} B and D). The improvement was modest for noise-free or high SNR data but was large for low SNR data and precision. The recall decreased from 90\% to 60\% with an increasing number of MUs. Also, the recall was approximately stable at >80\% for all noise levels.

\begin{figure*}[ht]
\begin{center}
\makebox[\textwidth]{\includegraphics[width=\textwidth]{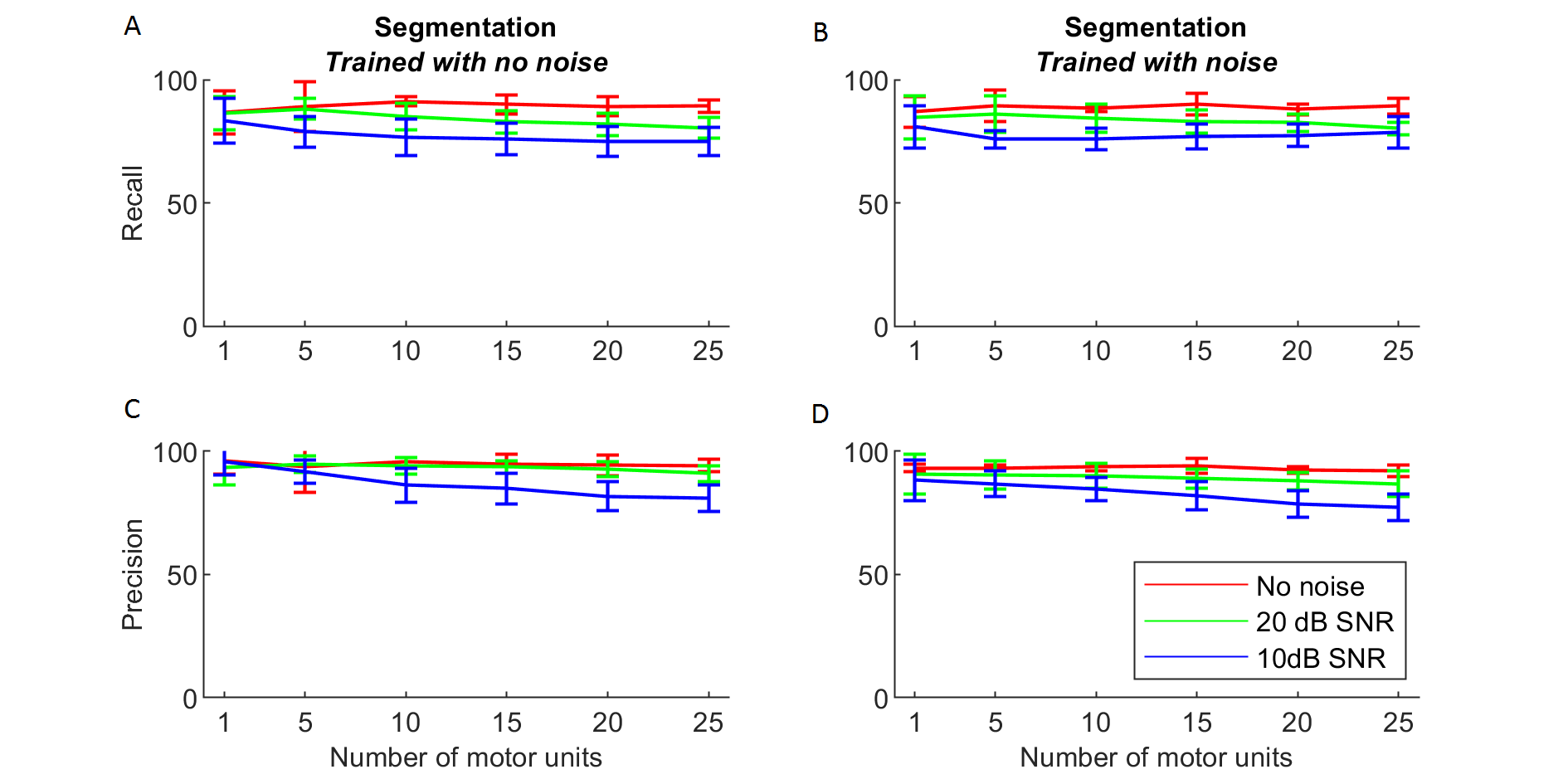}}
\end{center}
\caption{Object segmentation recall (A) and (B) and precision (C) and (D). The segmentation model results when evaluated on the test set with varying numbers of motor units and noise. Each data point is computed as the mean of 100 examples.   Figure \ref{fig:segment_results}(A) and (C) correspond to the model trained with noise-free data. Figure \ref{fig:segment_results} (B) and (D) correspond to the model trained on noisy data.}
\label{fig:segment_results}
\end{figure*}

The segmentation results are shown in Figure \ref{fig:segment_results} (A) and (C) for training with noise-free data and Figure \ref{fig:segment_results} (B) and (D) training with noisy data respectively. The number of MUs did not significantly influence the segmentation performance. Also, training with noise did not impact the segmentation performance compared to when it was trained on noise-free data.
%%%%%%%%%%%%%%%%%%%%%%%%
%SEGMENTATION RESULTS 
%%%%%%%%%%%%%
%%%%%%%%%%%%%

\subsection{Twitch train estimation performance}
\begin{figure*}[ht]
\begin{center}
\makebox[\textwidth]{\includegraphics[width=\textwidth]{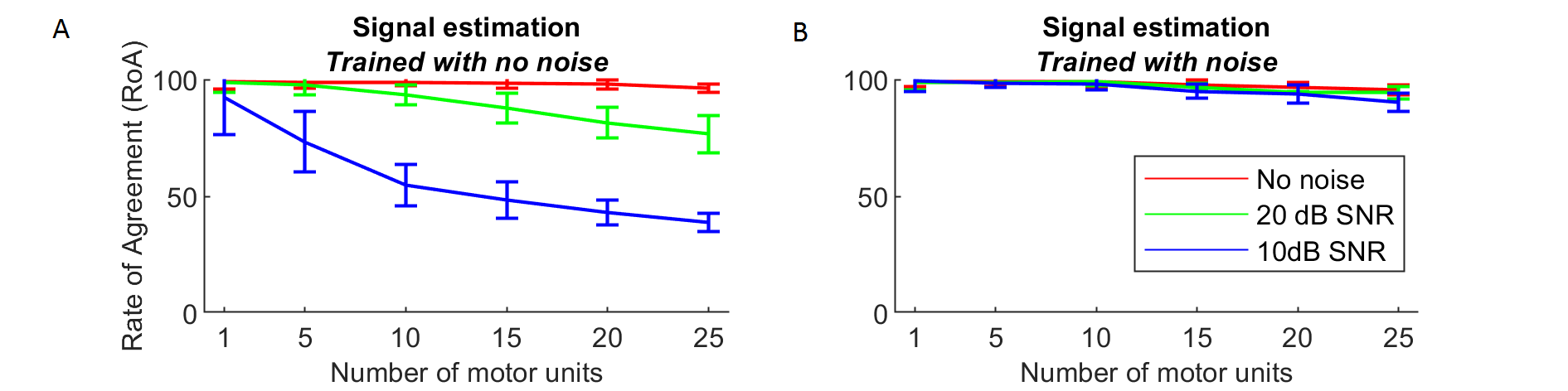}}
\end{center}
\caption{Rate of agreement of the twitch train estimation model for evaluation on different noise levels and a different number of motor units. (A) The model is trained with no noise data. (B) The model is trained with noisy data.}
\label{fig:roa_results}
\end{figure*}

\begin{figure*}[ht]
\label{Conv}
\begin{center}
\includegraphics[width=0.9\textwidth]{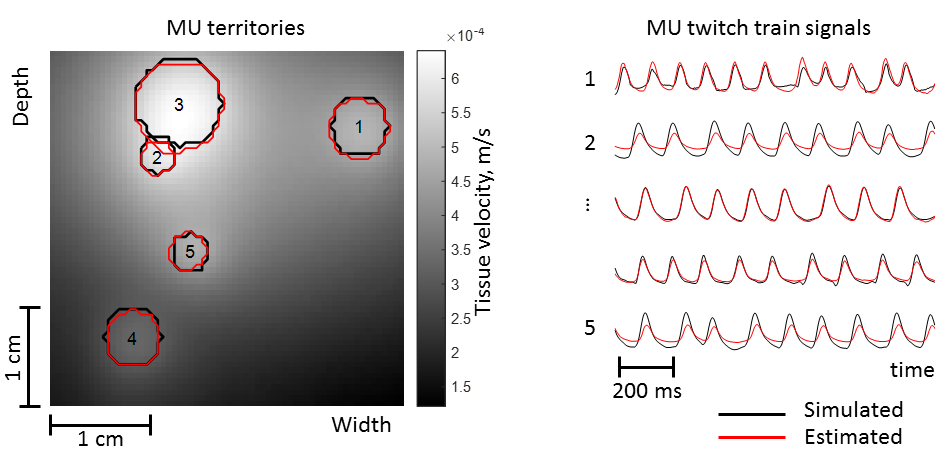}
\end{center}
\caption{Example of results from a simulated tissue velocity image sequence with five active MUs. The left panel shows the simulated and segmented MU territories and an underlying map of the tissue velocity distribution within the image. The right panel shows the simulated and estimated twitch train signals of the corresponding MUs.}
\label{fig:Example5MUsTrueAndEstimated}
\end{figure*}

\begin{figure*}[ht!]
\label{Conv}
\begin{center}
\includegraphics[width=0.9\textwidth]{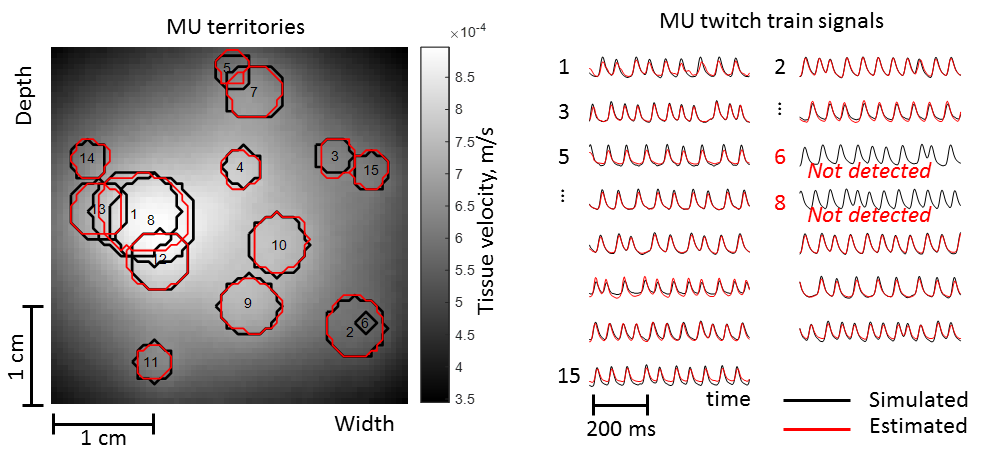}
\end{center}
\caption{Example of results from a simulated tissue velocity image sequence with 15 active MUs. The left panel shows the simulated and segmented MU territories and an underlying map of the tissue velocity magnitude distribution within the image. The right panel shows the simulated and estimated twitch train signals of the corresponding MUs.}
\label{fig:Example15MUsTrueAndEstimated}
\end{figure*}

The twitch train (signal) estimation model's performance can be seen in Figure \ref{fig:roa_results}. When trained without noise, the model performs almost perfectly for low noise case, but performance drops when noise is present, mainly when more MUs are present (Fig \ref{fig:roa_results} A). When the model was trained with noise, the performance was over 90\% independently on the number of active MUs or the data's noise level (Fig \ref{fig:roa_results} B). 

%%%%%%%%%%%%%%%%%%%%%%%%
%ESTIMATION RESULTS

%%%%%%%%%%%%%
%%%%%%%%%%%%%%%%%%%%%%%
%5 units
%%%%%%%%%%%%%%%%%%%%%%%%%%%%%
%16 units
%%%%%%%%%%%%%%%%%%%%%%%%%%%%%%%%%%%%%%%%%%%%%

\section{Discussion}
In this work, a deep learning pipeline is suggested to identify MUs, segmentation of their territories, and estimate their twitch train activation based on ultrasound image sequence data of skeletal muscle contractions. The proposed model's performance is evaluated using simulated ultrasound data mimicking the complex activation pattern of tens of activated MUs with overlapping territories and partially synchronized activation patterns. Performance evaluation shows that the proposed pipeline can effectively identify individual MUs, and estimate their territories and twitch train signal at low contraction forces.

\subsection{Evaluation of performance}
First, the influence of including noisy signals in training was large. When models were trained on noise-free data, the performance was significantly lower for noisy signals as compared to noise-free signals (e.g., Figure \ref{fig:detect_results} and Figure \ref{fig:roa_results}). When training the models with noisy signals, the performance was high independently of the noise level. These observations are consistent with the results in the literature on the robustness of CNNs on noisy images. The addition of noise during the deep learning architecture training has been argued to have a regularization effect and, thus, gives robustness to the model \cite{noiseeffect1996, bishop2008}. 

Second, the results showed that the detection recall decreased with the activation level. For N=25 units active, about 60\% of the MUs could be detected. The recall and precision of the territories' segmentation were greater than 80\%, and RoA was greater than 90\% for all activation levels (N=1 -- 25 MUs) when trained with noisy signals. Compared to the previous work by Rohl{\'e}n et al. \cite{rohlen2020method}, who used the same simulation method and evaluation metrics, the detection performance also decreased with increasing activation level. However, at 25 MUs, their method achieved a higher recall of 75\%. The proposed deep learning pipeline consistently outperformed Rohl{\'e}n et al. \cite{rohlen2020method} regarding the territory segmentation and firing pattern estimation, which had a recall in the range of 50 to 60\% and a declining pattern for RoA vs. activation level, achieving 60\% for 25 active MUs. In summary, we observe that the deep learning pipeline has better performance in terms of the segmentation and twitch-train estimation. However, the detection task needs further improvement.

\subsection{Limitations}
The deep learning pipeline was trained using simulated muscle contraction data \cite{rohlen2020method}. In this context, its performance should be interpreted as a proof-of-concept and demonstration of the principle. A limitation of this approach is that the method has learned the simulation data features compared to the previously suggested approach using blind source separation \cite{rohlen2020method}, which does not rely on learning the data is a more generalized approach.
Consequently, we do not expect that the present trained network should have necessarily high performance on experimental data. There are three key simplifications/differences of the simulated data compared to experimental data: 1) superposition of mechanical responses of multiple MUs, 2) no extracellular matrix (fascia) connecting the fibers is included, and 3) non-physiological noise is additive and white. The first assumption has been shown valid at low force contractions where only a limited number of MUs are active as in our case \cite{rohlen2020method}, \cite{yoshitake2002}, \cite{cescon2008}. The second assumption is a simplification of the true anatomy. Still, it has been indicated that in the cross-sectional view of the muscle, the spatio-temporal pattern is highly similar, comparing image sequences of simulated and experimental muscle contractions \cite{rohlen2020method}. 
The impact on the third assumption's training was clear in the results and shows that a relevant noise model will be influential on performance. 

A potential solution to translate the pipeline to an experimental application is to use the trained model from this work and transfer learning. This translation can be done by fine-tuning the model with labeled experimental data from gold-standard measurements of invasive EMG methods (needle-EMG) that can record the activation of individual MUs \cite{preston2012electromyography}.

From the computational perspective, implementing the two key modules is based on a number of deep learning architectures, essentially requiring GPUs to perform training and inference. So, it would be useful to build methods with lower computational complexity while attaining a comparable performance. We can certainly hope that given the rising popularity of deep learning methods for medical imaging in general, developing more computationally efficient methods for motor unit detection and twitch train estimation is only a matter of time. 

\subsection{Applications}
The ability to identify the activity of MUs in the whole muscle (large field of view) would allow larger accessibility than current EMG methods that suffer from a small field of view. The proposed technique has many interesting and important applications, given a successful translation training the pipeline on experimental data. For example, for recording the neural firing patterns of MUs to control prostheses \cite{farina2017man}, studying strategies of the central nervous system on MU recruitment in endurance/fatiguing tasks \cite{holtermann2008differential}, or clinical diagnostics when territories and/or firing pattern are altered due to pathological processes e.g., \cite{staalberg1990use}. Altogether, the proposed method could allow the study of various questions that previously were difficult or not possible to address.

\section{Conclusions}
In this work, a deep learning pipeline is suggested to identify the mechanical response of individual MUs, segmentation of their MU territories, and estimate their twitch train activations based on ultrasound image sequence data in voluntary skeletal muscle contractions.
The results show that the proposed pipeline can effectively identify individual MUs, and estimate their territories and twitch train signal at low contraction forces. The proposed method is potentially useful to progress with experimental data. The ability of an ultrasound imaging based non-invasive large field of view of the active MUs would make it possible to address a variety of questions that were difficult to address before.

\section*{Acknowledgment}
The authors thank M.D. Lars-Johan Liedholm at Dept of Neurophysiology, V{\"a}sterbotten County Council, Ume{\aa}, Sweden, for useful discussions in the initial phase of the work.

\bibliographystyle{IEEEtran}
\bibliography{paper}
%%%%%%%%%%%%%%%%%%%%%%%%%%%%%%%%%%%%%%%%%%%%%

\appendices
\section{Losses for segmentation model}
\label{appendixA}
\begin{figure}[h!]
\centering
  \begin{subfigure}[b]{0.45\textwidth}
    \includegraphics[width=\textwidth]{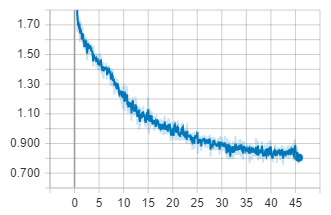}
    \caption{Training loss.}
    \label{fig:seg_train_loss}
  \end{subfigure}
  
  \begin{subfigure}[b]{0.45\textwidth}
    \includegraphics[width=\textwidth]{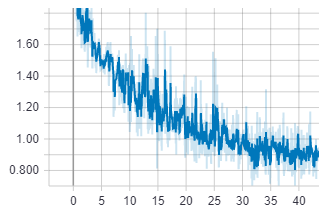}
    \caption{Test loss.}
    \label{fig:seg_test_loss}
  \end{subfigure}
  \caption{Losses plotted over time (hours) when training the segmentation model. Ten mini-batches of validation data are computed for every 100 mini-batches of training data.}
  \label{fig:seg_losses}
\end{figure}

Losses for the segmentation model on the training and validation sets are shown in Figure \ref{fig:seg_losses}.

\section{Losses for Time signal estimation model}
\label{appendixB}
\begin{figure}[h!]
  \begin{subfigure}[b]{0.45\textwidth}
    \includegraphics[width=\textwidth]{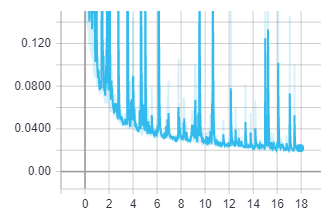}
    \caption{Training loss.}
    \label{fig:transc_train_loss}
  \end{subfigure}
  \begin{subfigure}[b]{0.45\textwidth}
    \includegraphics[width=\textwidth]{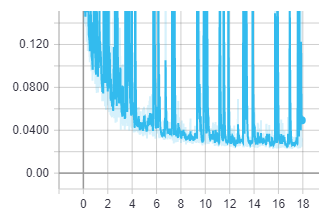}
    \caption{Validation loss.}
    \label{fig:transc_test_loss}
  \end{subfigure}
  \caption{Losses plotted over time (hours) when training the signal estimation model. Ten mini-batches of validation data is computed for every 100 mini-batches of training data.}
  \label{fig:transc_losses}
\end{figure}

Losses for the twitch train estimation model on the training and validation sets are shown in Figure \ref{fig:transc_losses}.
%%%%%%%%%%%%%%%%%%%%%%%%%%%%%%%%%%%%%%%%%%
\section{RoA Computation}
\label{appendixC}
\begin{figure*}[ht!]
\begin{center}
\makebox[\textwidth]{\includegraphics[width=\textwidth]{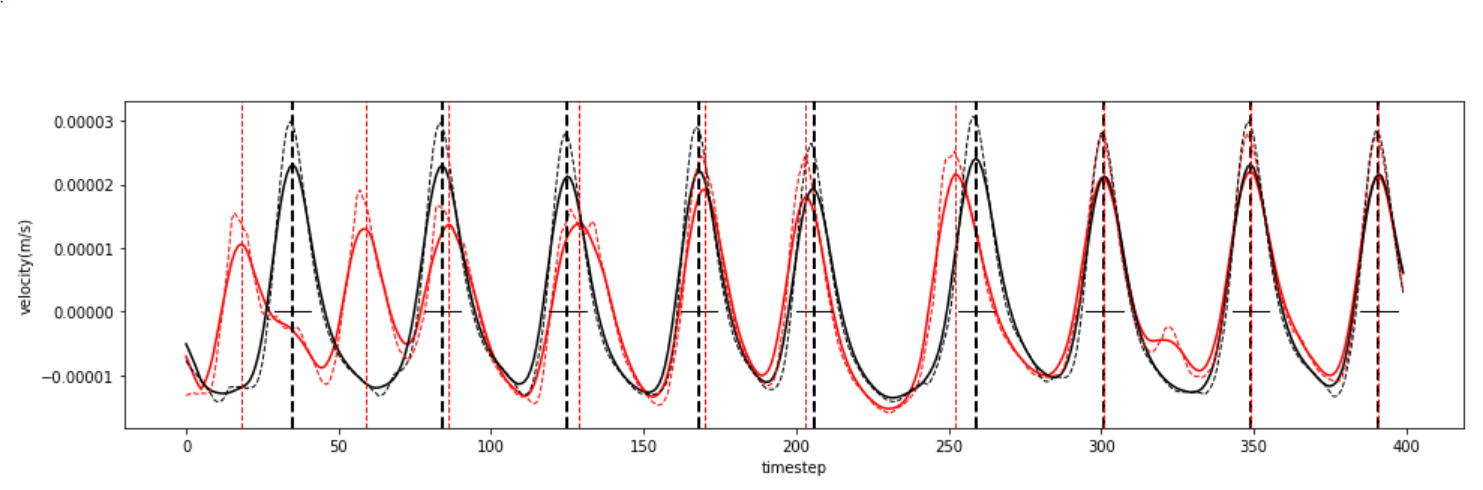}}
\end{center}
\caption{Illustration of an RoA computation. The red marks correspond to the predicted signal, and the black marks correspond to the simulated ground truth signal. The dashed curves are the original signals, and the solid curves are smoothed versions of those. Vertical lines are the estimated firings, and horizontal lines are the intervals within which firings are considered to match. In this example $c=7, A=3$ and $B=2$ resulting in $RoA=0.58\overline{3}$.}
\label{fig:RoA_tutorial}
\end{figure*}

The computation of rate of agreement is illustrated in Figure \ref{fig:RoA_tutorial}.
%%%%%%%%%%%%%%%%%%%%%%%%%%%%%%%%%%%%%%%%%%%%%%
\section{Examples of simulated image sequences}
\label{appendixD}
\begin{figure*}[ht!]
\label{Conv}
\begin{center}
\includegraphics[width=0.9\textwidth]{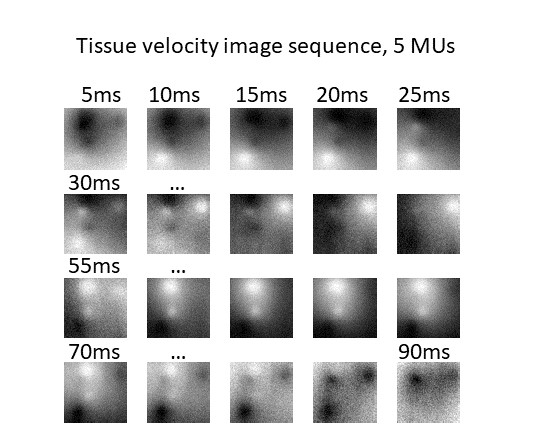}
\end{center}
\caption{Example of a simulated tissue velocity image sequence with five active MUs. Approximately one cycle of the contraction of the units can be seen. White color means contraction and dark color relaxation.}
\label{fig:ImageSequence5MUs}
\end{figure*}

\begin{figure*}[ht!]
\label{Conv}
\begin{center}
\includegraphics[width=0.9\textwidth]{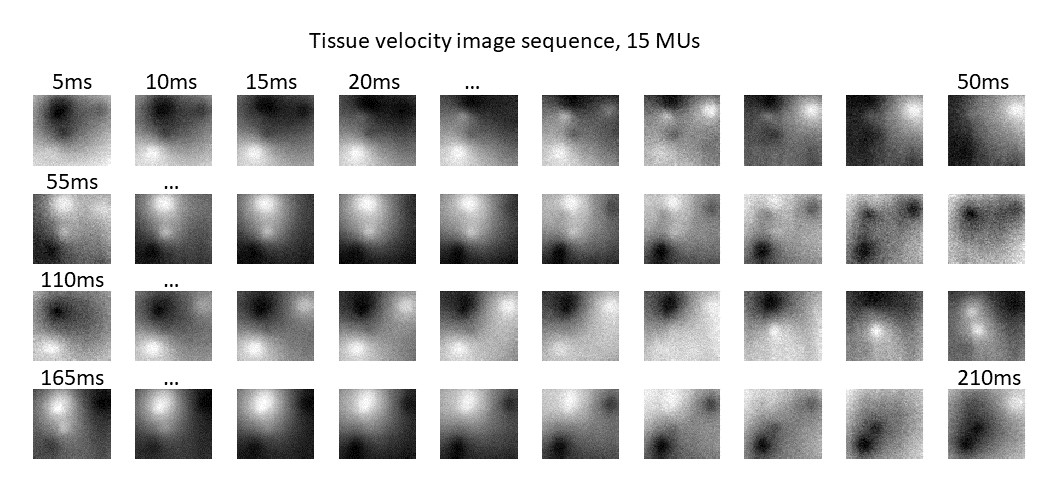}
\end{center}
\caption{Example of a simulated tissue velocity image sequence with 15 active MUs. Approximately two cycles of the contraction of the units can be seen. White color means contraction and dark color relaxation.}
\label{fig:ImageSequence15MUs}
\end{figure*}

Examples of simulated tissue velocity image sequences for 5 MUs and 15 MUs are shown in Figure \ref{fig:ImageSequence5MUs} and Figure \ref{fig:ImageSequence15MUs}, respectively. 

%%%%%%%%%%%%%%%%%%%%%%%%%%%%%%%%%%%%%%%%%%%%%
%%%%%%%%%%%%%%%%%%%%%%%%%%%%%%%%%%%%%%%%%%%%%

\end{document}